\documentclass{article}

\usepackage{microtype}
\usepackage{graphicx}
\usepackage{subfigure}
\usepackage{booktabs} 

\usepackage{hyperref}

\usepackage[accepted]{icml2024}

\usepackage{amsmath}
\usepackage{amssymb}
\usepackage{mathtools}
\usepackage{amsthm}

\usepackage[capitalize,noabbrev]{cleveref}

\theoremstyle{plain}

\theoremstyle{definition}

\theoremstyle{remark}

\usepackage[textsize=tiny]{todonotes}

\usepackage{booktabs}
\usepackage{paralist}
\usepackage{subcaption} 
\usepackage{amsmath}
\usepackage{amsfonts}       
\usepackage{mathtools}
\usepackage{bm}
\usepackage{dsfont}

\newcommand{\D}{\mathcal{D}}
\newcommand{\F}{\mathcal{F}}
\newcommand{\G}{\bm{\theta}_\D} 
\newcommand{\R}{\bm{\phi}_\D}   
\newcommand{\n}{n_\rho^f}

\newcommand{\s}[1]{\mathop{size}(#1)}

\icmltitlerunning{Bayesian Program Learning by Decompiling Amortized Knowledge}

\begin{document}

\twocolumn[
\icmltitle{Bayesian Program Learning by Decompiling Amortized Knowledge}



\icmlsetsymbol{equal}{†}

\begin{icmlauthorlist}
\icmlauthor{Alessandro B. Palmarini}{sfi,equal}
\icmlauthor{Christopher G. Lucas}{uoe}
\icmlauthor{N. Siddharth}{uoe,ati}
\end{icmlauthorlist}

\icmlaffiliation{sfi}{Santa Fe Institute, Santa Fe, NM, USA}
\icmlaffiliation{uoe}{School of Informatics, University of Edinburgh, Edinburgh, UK}
\icmlaffiliation{ati}{The Alan Turing Institute, UK}

\icmlcorrespondingauthor{Alessandro B. Palmarini}{abp@santafe.edu}

\icmlkeywords{ICML, library learning, program synthesis, program induction, Bayesian program learning, wake-sleep, amortized inference, DreamCoder}

\vskip 0.3in
]



\newcommand{\ugnotice}{\textsuperscript{†}Majority of work done at University of Edinburgh}
\printAffiliationsAndNotice{\ugnotice}

\begin{abstract} 
    %
    \textsc{DreamCoder} is an inductive program synthesis system that, whilst solving problems, learns to simplify search in an iterative wake-sleep procedure.
    The cost of search is amortized by training a neural search policy, reducing search breadth and effectively ``compiling'' useful information to compose program solutions across tasks. 
    Additionally, a library of program components is learnt to compress and express discovered solutions in fewer components, reducing search depth.
    %
    %
    We present a novel approach for library learning that directly leverages the neural search policy, effectively ``decompiling'' its amortized knowledge to extract relevant program components.  
    This provides stronger amortized inference: the amortized knowledge learnt to reduce search breadth is now also used to reduce search depth.
    We integrate our approach with \textsc{DreamCoder} and demonstrate faster domain proficiency with improved generalization on a range of domains, particularly when fewer example solutions are available.
\end{abstract}

\section{Introduction}
\label{sec:introduction}



The goal in inductive program synthesis is to generate a program whose functionality matches example behaviour \cite{gulwani2017program}. If behaviour is specified as input/output examples, then the problem could be solved trivially by defining a program that embeds (hard codes) the example transformations directly. This solution, however, does not provide an account relating, or explaining, how inputs map to outputs, nor does it facilitate generalization to inputs beyond what it has observed. Discovering a program that can do this cannot be derived from examples, just as our scientific theories cannot be derived from observations \cite{deutsch2011beginning}---some form of hypothesising is required (i.e. search).

Any program must be built from some set of primitive components: base operations and elements. While a set of such components (library) is finite, they can be combined in an infinite number of ways to form different programs. Moreover, this number grows exponentially with the number of components used, rendering blind search intractable in all but the simplest of cases.
%
Improving upon blind search requires reducing the number of combinations searched. 
There are two broad routes to achieving this: we can (i) reduce the \emph{breadth} searched as we move down the search tree to program solutions or we can (ii) reduce the \emph{depth} searched by pushing program solutions higher up the search tree. Accomplishing either requires additional \emph{knowledge}.

\begin{figure}[t]
    \centering
    \includegraphics[width=0.85\columnwidth]{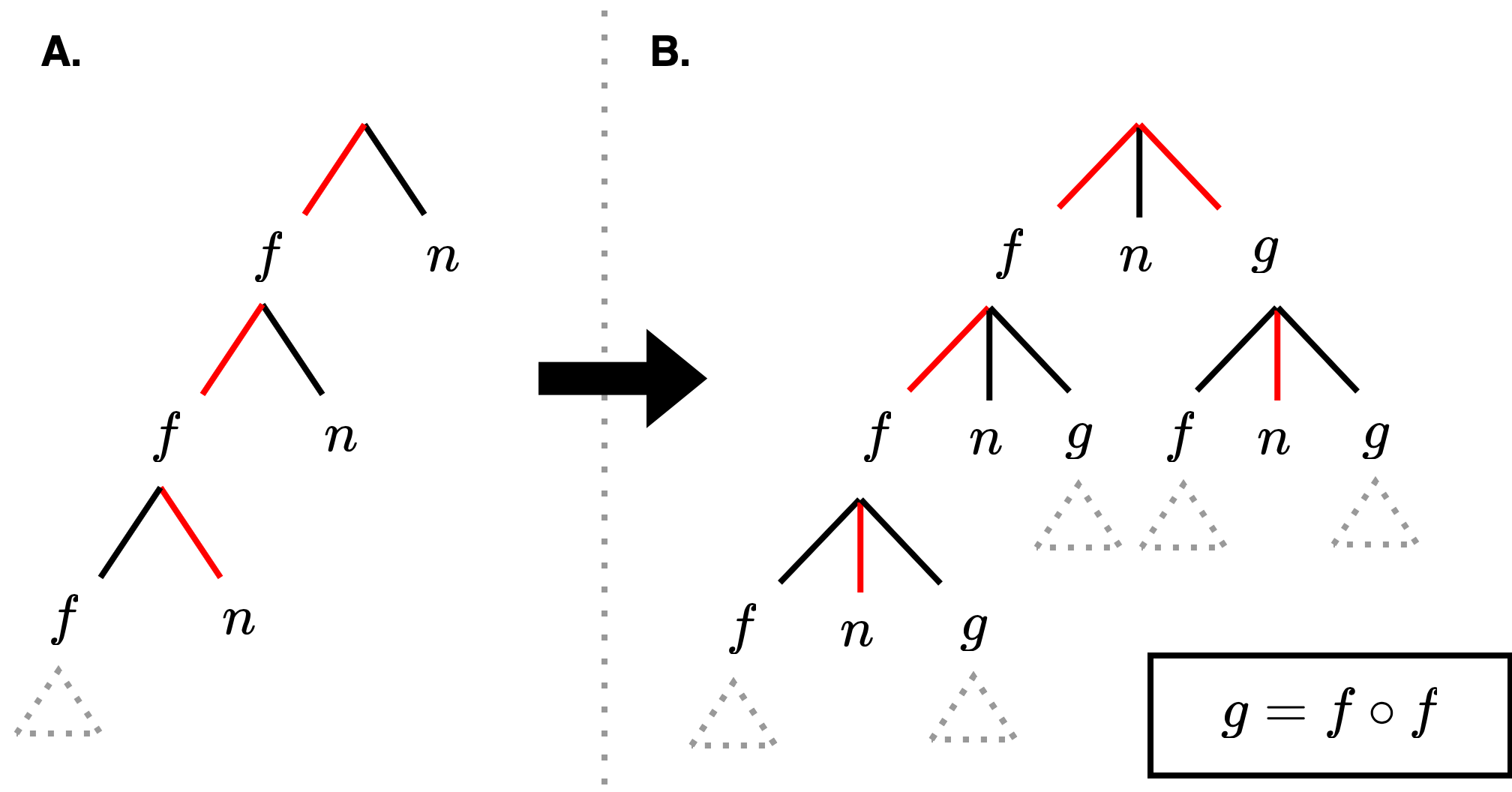}
    \caption{\textbf{(A)} Program search space for a simple library containing two primitives: a function \(f : \mathbb{N} \rightarrow \mathbb{N}\) and a variable \(n : \mathbb{N}\). Red path highlights the program \(f(f(n))\). \textbf{(B)} Restructured search space after \(f \circ f\) is chunked into a new primitive \(g\), illustrating the breadth-depth trade off.} 
  \label{fig:search-tree-chunk}
\end{figure}

To reduce the breadth searched requires knowledge for avoiding search paths that do not contain program solutions, while focusing on those that do---i.e. a search policy.
To reduce the depth searched requires knowledge about which functionality is used by program solutions: if any functionality, expressed as a combination of library components, was itself made part of the library, then any program using this functionality could be expressed with fewer components.
In effect, a copy of every program containing the functionality is pasted higher up the search tree, reducing the depth required to reach the program's corresponding computation, as illustrated in Fig. \ref{fig:search-tree-chunk}. We refer to turning composed functionality into a library operation as ``chunking''.

Chunking to reduce the search depth comes at the cost of increasing the search breadth---there are now more library components to consider (Fig. \ref{fig:search-tree-chunk}). Hence, \emph{the knowledge for successfully reducing the depth searched in a program induction task is highly reliant on the knowledge used to reduce the breadth}, and vice versa. This interplay is key to the main problem addressed in this paper. 


Practical and successful program synthesis approaches, such as \textit{FlashFill} \cite{gulwani2011automating}, involve the careful selection of its library components, i.e., the domain-specific language (DSL) used.
Human experts define a search space where programs capable of solving specific tasks can be found tractably, but these systems are typically unable to solve tasks not anticipated by the designers---failing to generalize.
A general-purpose inductive program synthesis system, then, cannot rely on handcrafted DSLs; instead, it must learn how to structure its search space, via the components contained in its library \cite{dechter2013bootstrap, dumancic2021knowledge, shin2019program, iyer2019learning}, and learn how to navigate that search space for program solutions \cite{balog2016deepcoder, devlin2017robustfill, kalyan2018neural, chen2018execution}.

\textsc{DreamCoder} \cite{ellis2021dreamcoder, ellis2023dreamcoder} is an inductive program synthesis system that jointly learns a library and a probabilistic search policy, but without explicitly considering the interplay between search depth and breadth described above. That is, the knowledge learnt to guide the search space has no direct effect on how it is restructured (i.e., how the library is learnt). We contribute a novel approach for library learning that directly leverages the existing knowledge learnt to guide search, enabling the extraction of more useful, and complementary, components from the same experience. 
We integrate our approach with \textsc{DreamCoder} and demonstrate faster domain proficiency with improved generalization on a range of program synthesis domains.

\section{Background}
\label{sec:dreamcoder}

We begin with an overview of \textsc{DreamCoder}'s approach: viewing program induction as probabilistic inference \cite{lake_human-level_2015}.
%
Consider a set of program induction tasks \(\mathcal{X}\). We assume each task \(x \in X\) was produced by an unobserved (latent) program \(\rho\) and each \(\rho\) was generated by some prior distribution \(p_{\G^*}(\rho)\). We assume the prior is defined by model parameters \(\G\) that specify the probability that components part of a library \(\D\) are used when generating programs \cite{liang2010learning, menon2013machine}. The true library and component parameters \(\G^*\) are unknown.

The marginal likelihood of the observed tasks is then given by \(p_{\G}(\mathcal{X}) = \prod_{x \in \mathcal{X}}\sum_\rho p(x \mid \rho) p_{\G}(\rho)\), where \(p(x \mid \rho)\) is the likelihood of \(x\) being produced, and hence solved, by \(\rho\). To learn a good generative model that maximises the likelihood, we need to know which programs score highly under \(p(x \mid \rho)\)---i.e. solve our tasks.
We have seen previously why this is challenging: discovering programs that can account for the observed tasks requires search.

To help with search, a recognition (inference) model \(q_{\R}(\rho \mid x)\) is learnt to infer the programs that are most likely to solve a given task.
The recognition model parameters \(\R\) map tasks to distributions over programs that, as with the generative model, specifies the probability that components part of the library \(\D\) are used.
Estimating \(\R\) is done using both \((\rho, x)\) pairs sampled from the generative model (fantasies) and programs found to solve the observed tasks \(x \in \mathcal{X}\) (replays).
The recognition model is used to search for programs that solve a task~\(x\) by enumerating programs in decreasing order of their probability under \(q_{\R}(\rho \mid x)\). Programs that solve \(x\) are stored in a task-specific set \(\mathcal{B}_x\).

%
Discovering programs that may have produced the observed tasks (those in \(\{\mathcal{B}_x\}_{x \in \mathcal{X}}\)) now provides more data to infer the parameters \(\G\) generating them. Inferring \(\G\) entails choosing the library \(\D\) whose components they control. 
Rather than maximise the likelihood directly, \textsc{DreamCoder} performs maximum a posteriori (MAP) inference using a prior over libraries \(\D\) and parameters \(\G\). 
Maximising the MAP objective (which can only be approximated) w.r.t. \(\D\) corresponds to updating \(\D\) to include functions that best compress the discovered solutions. After updating \(\D\), parameters \(\G\) are updated to their MAP estimates.



This completes a single learning cycle with \textsc{DreamCoder}---a variant of the wake-sleep algorithm \cite{hinton1995wake, bornschein2014reweighted, le2020revisiting}. An iterative procedure is used to jointly learn a generative and inference model \cite{dayan1995helmholtz}, while additionally searching for program solutions.
%
After updating the generative model, it can then be used
to sample \((\rho, x)\) pairs more indicative of those coming from the true generating process. These can be used to learn a more accurate recognition model with improved inference (dream sleep), which results in more programs capable of solving tasks discovered during search (wake). More program solutions lead to better estimates of the generative model (abstraction sleep). Each stage bootstraps off that learnt in the previous stage.



\begin{figure}[ht]
    \centering
    \includegraphics[width=0.85\columnwidth]{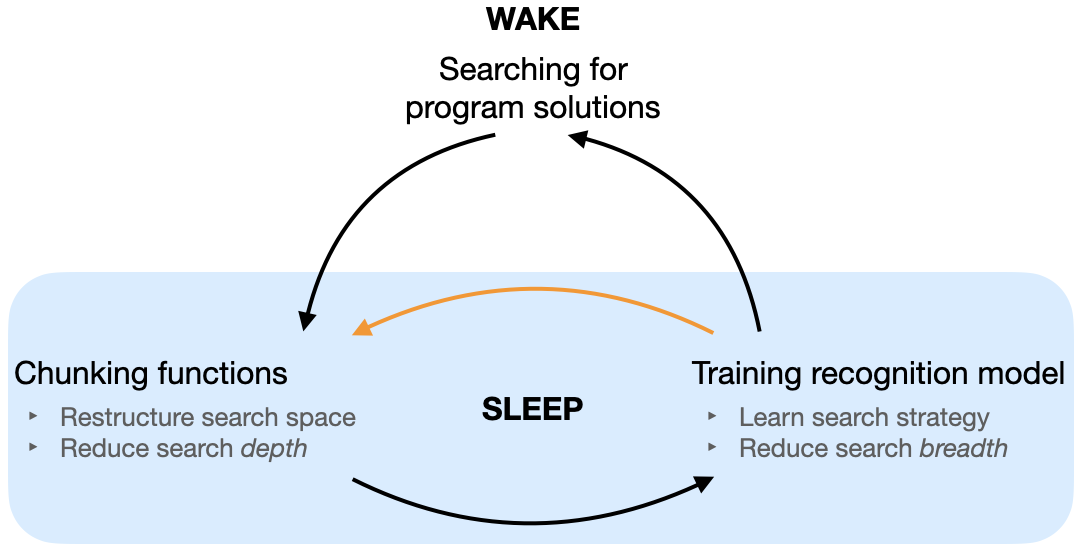}
    \caption{Positive feedback loop between searching for program solutions and learning to simplify search in \textsc{DreamCoder} (black arrows). In dream decompiling (orange arrow) the knowledge learnt to reduce search breath directly influences the knowledge learnt to reduce search depth.}
    \label{fig:chain-of-influence}
\end{figure}

\section{Dream Decompiling}
\label{sec:ddc}

Learning a library \(\D\) and recognition model parameters \(\R\) during \textsc{DreamCoder}'s two sleep phases corresponds to learning the two types of knowledge (discussed in Section \ref{sec:introduction}) to simplify program search (Fig.~\ref{fig:chain-of-influence} black arrows). Adding functions to \(\D\) reduces the search depth to reach programs utilizing those functions. Searching for programs under \(q_{\R}(\rho \mid x)\) directs the search down preferred branches. Chunking (to reduce search depth) restructures the search space that the recognition model guides (to reduce search breadth). 
How the search space is restructured (here by adding to \(\D\)) should complement the search policy (here \(q_{\R}(\rho \mid x)\)) guiding that space. For example, consider a task with multiple solutions: reducing the depth to reach some of those solutions does not simplify search if the guide is focused on others. In \textsc{DreamCoder}, the recognition model has only an indirect effect on what is chunked through the program solutions it helps discover (as they then determine which functions are most compressive).
This raises the question: can we improve inference by \emph{directly} leveraging the knowledge learnt by the recognition model to decide what is chunked, and hence have the recognition model shape the search space that it will subsequently learn to guide? 


The recognition model uses one set of parameters (\(\R\)) to model the relation between observed tasks and latent programs that may have produced those tasks. A single function (parameterised by \(\R\)) is used to specify an inference distribution \(q(\rho \mid x)\) for any \(x\), rather than learning each individually. This is called amortized inference \cite{gershman2014amortized, stuhlmuller2013learning}: search on new tasks becomes tractable by reusing knowledge learnt from past inferences. 
By leveraging the recognition model to decide what to chunk, we now amortize the cost of search (inference) in two ways: recognition model parameters \(\R\) are reused to reduce both search breadth \textit{and} depth.

The amortized model learnt by the recognition model offers the potential to chunk valuable functions for \textit{unsolved} tasks, due to its ability to generalize. Unsolved tasks have no example solutions to infer compressive functionality. Moreover, the functionality crucial for solving new tasks may not be present or sufficiently abundant in existing solutions to other tasks. Such functionality is thus overlooked by compression-focused approaches to library learning: dealing with future uncertainty is often antagonistic to optimal compression of what has worked in the past \cite{chollet2019measure}.



Training the recognition model on \((\rho, x)\) pairs ``compiles'' useful information into model parameters \(\R\) for generating program solutions across tasks \cite{le2017inference}.
The fantasy \((\rho, x)\) pairs sampled from the generative model are commonly referred to as ``dreams'' \cite{pml2Book}.
Chunking with the recognition model (Fig.~\ref{fig:chain-of-influence} orange arrow) involves utilizing the compiled information to identify the most useful functions to incorporate into the library, essentially \emph{decompiling} the knowledge acquired through amortized inference. This is analogous to a decompiler translating compiled machine code into high-level source code. Hence, we refer to this concept as ``dream decompiling''.

\section{Approach}
\label{Section: Approach}

In this section we present two variants of dream decompiling, addressing the problem of choosing effective library components.
For simplicity, we consider the case where the recognition model defines a probability distribution over library components and that the probability of generating a program is equal to the product of independently generating each component of the program. Appendix \ref{A:flexible_q} describes how this extends to bigram distributions over well-typed programs (as is the case with \textsc{DreamCoder}).

A simple approach to chunking with the recognition model is to consider which functions it wants to use most often, irrespective of the particular task being solved:
\begin{equation}
    \footnotesize
    q_{\R}(f) = \mathbb{E}_{x \sim \mathcal{X}}[q_{\R}(f \mid x)]
    \label{Eq: simple approach}
\end{equation}
We refer to this variant as \textsc{DreamDecompiler-Avg}.
While this provides a means to rank each \(f\) in a set of candidates \(\F\), there are two problems with this approach:
\begin{compactenum}[i.]
    \item A ranking can determine if chunking one function is better than another, but it provides no insight into how many functions (if any) should be chunked.

    \item Larger functions are naturally harder to generate than smaller functions and hence disfavoured in the ranking. To see why this can be problematic, consider a strict subfunction $s$ of a function $f$. On all tasks, the recognition model is less likely to generate $f$ than $s$ as it would need to generate all components in $s$ plus more. But if the recognition model is only ever intending to generate $s$ as part of $f$, then chunking $f$ is better: future use would require generating a single component instead of multiple and $s$ has no use elsewhere.
\end{compactenum}

If we want to leverage the recognition model for chunking, we need a criterion that (i) determines which functions should actually be chunked and (ii) considers the overall impact of chunking a function, not only generation preference.

It is unknown whether chunking a function will enhance, diminish, or have no impact on the recognition model's ability to guide the search for program solutions.
A formal way to handle uncertain knowledge is with probability distributions \cite{cox1946probability}. In this section we introduce a probabilistic model to systematically express the uncertainty in whether or not chunking a given function will have an overall beneficial effect on the recognition model's future ability to guide search for program solutions. The model, written \(p(c \mid f; \R)\), is a Bernoulli distribution over the binary random variable \(c\) that, w.r.t. some function \(f\), has the value \(1\) if the net effect of chunking \(f\) is positive and \(0\) otherwise. 
%
%
The distribution used depends on the recognition model \(q_{\R}(\rho \mid x)\) and is hence parameterised by \(\R\).

With the model \(p(c \mid f; \R)\) we can solve both of the above problems. The first can be solved using decision theory \cite{degroot2005optimal}. For example, if the utility gained from a favorable chunk was equal to the cost incurred from an unfavorable chunk, then opting to chunk a function \(f\) only when \(p(c \mid f; \R) \ge 0.5 \) would maximise the expected payoff.\footnote{This is only true for symmetric payoffs. Exploring payoffs and their assignment is left for future work. Here, the simplest quantity selection suffices to showcase the strengths of the approach.}
%
The extent to which the second problem is solved depends on how well the true distribution is modeled by \(p(c \mid f; \R)\), which we now explain how to do effectively.\footnote{Note that calculating the exact value of \(c\) with respect to all future tasks is not possible, and hence one can only have a heuristic.}


A model of \(p(c \mid f; \R)\) would require balancing the effect of chunking on the recognition model's future ability to generate all desired programs, across all tasks. The problem can be simplified by introducing and then marginalising out the programs and tasks instead:
\begin{equation}
    \footnotesize
    p(c \mid f; \R) = \sum_{x} \sum_{\rho} p(c \mid f, \rho, x; \R) p(\rho, x \mid f; \R)
    \label{Eq: double summation}
\end{equation}
We will return to the double summation and joint \(p(\rho, x \mid f; \R)\) later. The problem is now reduced to needing a model, \(p(c \mid f, \rho, x; \R)\), expressing uncertainty in whether or not chunking \(f\) will have a positive effect on the recognition model's ability to generate \(\rho\) for solving \(x\).

Note that knowing \(f\) is part of \(\rho\) (which can solve \(x\)) does not guarantee a positive effect from chunking \(f\). After \(f\) is added to the library, the recognition model parameters \(\R\) (dependent on \(\D\)) are updated, defining a new distribution over programs in terms of the new library. The effect of chunking \(f\) is a \emph{balance} between the improvement gained in generating \(f\) and any costs incurred in generating the rest of \(\rho\). We can not know the exact effect of chunking as we do not know how \(\R\) will be updated.

A related effect that (i) does not depend on how \(\R\) will be updated, (ii) can be captured precisely and (iii) calculated efficiently is how beneficial is it for the recognition model to generate \(\rho\) if it does not need to generate \(f\) at all---which we refer to as ``caching'' \(f\). In terms of the balance mentioned above, this is equivalent to the optimistic view that chunking \(f\) will result in maximal improvement for generating \(f\) (it can be generated for ``free''), while also not incurring any cost for generating the rest of \(\rho\). 

\label{sec:desiderata}

\textbf{Desiderata for a measure of caching benefit}~~
Let us consider a program \(\rho\) generated by a distribution \(q(\rho)\) via the composition of smaller library components. We are interested in a mathematical way to quantify the benefit gained, from \(0\) (not useful at all) to \(1\) (as useful as can be), in generating \(\rho\) if instead the function \(f\) was cached (no longer required to be sampled), which we denote \(\mathbb{C}(q, \rho, f)\).
Any such measure should satisfy the following properties:
\begin{compactenum}[\bfseries D1:]
  \item (\emph{min}) If \(f\) is not utilized by \(\rho\), then caching \(f\) provides no benefit and \(\mathbb{C}(q, \rho, f)\) should be \(0\).
  \item (\emph{max}) The largest benefit is gained from caching the entire program (no matter how \(q\) is defined). If \(f = \rho\), then \(\mathbb{C}(q, \rho, f)\) should be \(1\).
  \item (\emph{monotonic---changing programs}) Consider a fixed function \(f\) used in different \(\rho\). 
  The less likely \(q\) is to generate the parts of \(\rho\) that are not part of \(f\), the smaller \(\mathbb{C}(q, \rho, f)\) should be because caching \(f\) has a smaller impact on helping \(\rho\) to be subsequently generated. As an extreme example, if the probability of generating the rest of our program became \(0\), then caching \(f\) does not help with then generating the full program at all.
  \item (\emph{monotonic---changing functions}) Consider different functions \(f\) used within the same initial program. The smaller \(q(f)\) is, the better it is to cache (not needing to generate) \(f\) and hence the larger \(\mathbb{C}(q, \rho, f)\) should be.
\end{compactenum}

\begin{figure}[h]
    \begin{center}
        \includegraphics[width=0.90\columnwidth]{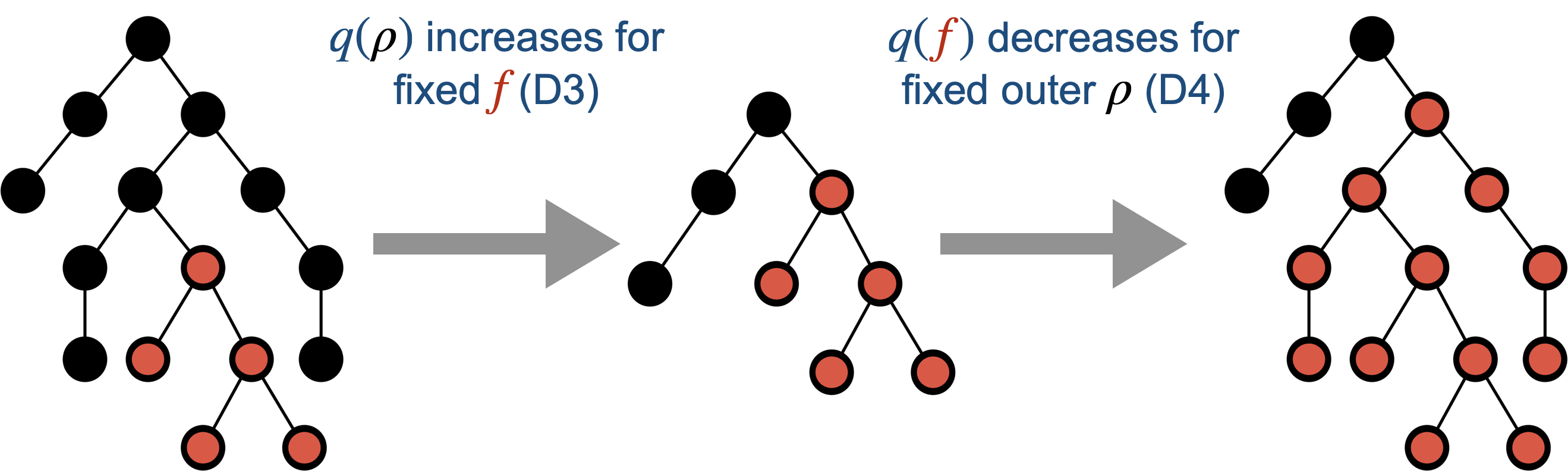}
        \caption{
        Program trees with functions (red subtrees) considered for caching.
        The probability of being generated by \(q\) decreases with an increase in nodes. Arrows indicate changes where caching becomes more beneficial.
        Caching the function in the first program is unhelpful, as the program remains difficult to generate. Caching the same function in the second program is more helpful, transitioning the program from relatively unlikely to likely. Caching the function in the third program is even more helpful: although post-cache sample probability is equal to the second, generating the third program was initially much more unlikely.
        %
        }
        \label{fig:properties}
    \end{center}
\end{figure}

The first two properties are straightforward to understand. The last two properties are illustrated further in Fig. \ref{fig:properties}.

A direct consequence of the last two properties is that the more components of a program cached as part of the function, the more beneficial the caching is. The last two properties are two sides of the same coin: both suggest that the benefit from caching \(f\) is related to the \textit{relative proportion of sample difficulty removed} when generating \(\rho\). In information-theoretic terms, ``difficulty'' can be used interchangeably with the ``self-information'' or ``surprise'' inherent in a random outcome (such as a function being generated) \cite{mackay2003information}. Capturing the self-information of a random outcome formally can be done with its negative log probability \cite{shannon1948mathematical}, leading to the following ratio:
\begin{equation}
    \footnotesize
    \mathbb{C}(q, \rho, f) = \frac{\n \log q(f)}{\log q(\rho)}
    \label{Eq:beneficial-caching-measure}
\end{equation}
where \(\n\) counts the number of times \(f\) appears in \(\rho\).

It is easy to verify that this definition satisfies each of our desired properties. 

\begin{compactenum}[\bfseries D1:]
  \item If \(f\) is not utilized by \(\rho\), then \(\n = 0\). Thus, \(\mathbb{C}(q, \rho, f) = \frac{0 \cdot  \log q(f)}{\log q(\rho)} = 0\).
  \item If the entire program is cached, then \(f = \rho\) and \(\n = 1\). Thus, \(\mathbb{C}(q, \rho, f) = \frac{1 \cdot  \log q(\rho)}{\log q(\rho)} = 1\).
  \item Consider the case where \(\rho\) contains one instance of \(f\). Splitting \(\rho\) into the components constituting \(f\) and those without (denoted \(\rho_{\setminus f}\)), we have that \(q(\rho) = q(f)q(\rho_{\setminus f})\) and
\begin{equation}
    \footnotesize
    \mathbb{C}(q, \rho, f) = \frac{\log q(f)}{\log q(f) + \log q(\rho_{\setminus f})}
    \label{Eq: split caching measure}
\end{equation}
As \(q(\rho_{\setminus f})\) decreases, the \textit{magnitude} of the denominator increases and hence \(\mathbb{C}(q, \rho, f)\) decreases.
  \item From Eq.~\eqref{Eq: split caching measure}: as \(q(f)\) decreases, \(\log q(\rho_{\setminus f})\) contributes less, and hence \(\mathbb{C}(q, \rho, f)\) increases.
\end{compactenum}

The measure of the benefit in caching can be used as a tractable estimate of the benefit in chunking and hence to define our probabilistic model of \(p(c \mid f, \rho, x; \R)\). Additionally, recall that \(p(c \mid f, \rho, x; \R)\) should express the uncertainty in whether or not chunking \(f\) will have a positive effect on the recognition model's ability to generate \(\rho\) for \emph{solving} \(x\). If \(\rho\) does not solve \(x\), then any benefit gained from chunking \(f\) to generate \(\rho\) does not matter:
\begin{equation}
    p(c = 1 \mid f, \rho, x; \R) = \mathds{1}[\rho \Rightarrow x]\mathbb{C}(q_{\R}, \rho, f)
\end{equation}
where \(\mathds{1}[\rho \Rightarrow x]\) equals \(1\) if \(\rho\) solves \(x\) and \(0\) otherwise.
As \(c\) is a binary random variable and \(0 \le \mathbb{C}(q_{\R}, \rho, f) \le 1\), the model expresses a valid probability distribution.

\label{Dealing with impossible summations}
The expression for \(p(c \mid f; \R)\) in Eq. \eqref{Eq: double summation} involves two impossible summations over the infinite sets of tasks and programs. We use a different approach to deal with each.

\textbf{Marginal over tasks}~~
Expressing the joint \(p(\rho, x \mid f; \R)\) as \(p(\rho \mid f, x; \R)p(x \mid f; \R)\) and then substituting \(p(x \mid f; \R)\) with its expression from Bayes' rule (which is proportional to \(p(f \mid x; \R)p(x; \R) = q_{\R}(f \mid x)p(x)\)) allows us to incorporate the recognition model's probability of generating \(f\) and, when substituted into Eq. (\ref{Eq: double summation}), express the marginal over tasks as an expectation. The model \(p(c \mid f; \R)\) thus becomes proportional to:
\begin{equation}
    \footnotesize
    \mathbb{E}_{p(x)}[q_{\R}(f \mid x) \sum_\rho p(c \mid f, \rho, x; \R) p(\rho \mid f, x; \R)]
    \label{Eq: marginal over programs}
\end{equation}
%
%
We can now form Monte Carlo estimates of the expectation using the observed tasks \(\mathcal{X}\). The normaliser  (equal to \(\mathbb{E}_{p(x)}[q_{\R}(f \mid x)]\)) can be estimated in the same manner.

\textbf{Marginal over programs}~~
We follow the approach taken by \citet{ellis2021dreamcoder} and marginalise over the finite beam of programs \(\mathcal{B}_x\) maintained for each task instead. This creates a lower bound particle-based approximation. Given that only the programs that solve \(x\) contribute probability mass to the summation, excluding any programs that do not solve \(x\) has no effect on the approximation. The same is true for programs that do not utilize \(f\). While all \(\rho \in \mathcal{B}_x\) solve \(x\), they do not necessarily contain \(f\). We can improve the approximation by attempting to refactor each \(\rho \in \mathcal{B}_x\) to a behaviourally equivalent program that instead contains \(f\). 
To deal with \(p(\rho \mid f, x; \R)\) in Eq. (\ref{Eq: marginal over programs}), we assume independence from \(f\), replacing it with \(q_{\R}(\rho \mid x)\).


\textbf{Final model}~~
We can understand $p(c \mid f; \R)$, our model expressing the uncertainty of whether chunking a function will benefit the recognition model's inference capabilities, as an interaction between three key sub-expressions:

\begin{equation}
    \footnotesize
     \mathbb E_{p(x)}[\underbrace{ q_{\R}(f \mid x)}_\textbf{(1)} \sum_{\rho \in \mathcal B_x} \underbrace{ \mathds{1}[\rho \Rightarrow x]\frac {n^f_{\rho}\log q_{\R}(f \mid x)} {\log q_{\R}(\rho \mid x)}}_\textbf{(2)} \underbrace{q_{\R}(\rho \mid x)}_\textbf{(3)}]
    \label{Eq:final-model}
\end{equation}

The functions with a high probability of being worthwhile to chunk are those that \textbf{(1)} first and foremost the recognition model \textit{wants} to generate as part of programs solving some task. When it does, \textbf{(2)} chunking the function greatly reduces the uncertainty (or we could say ``difficulty'') in \textbf{(3)} generating the recognition model's preferred programs to solve that task.
We refer to this probabilistic chunking variant of dream decompiling as \textsc{DreamDecompiler-PC}. 

\section{Case Analysis: Chunking With a Uniform q} 

In this section we examine the use of \(p(c \mid f; \R)\) as a chunking criterion when the recognition model distribution, parameterised by \(\R\), is uniform.
Here, the probability of generating a function \(f\) is \(1/|\D|^{\s{f}}\). The beneficial caching measure of Eq. (\ref{Eq:beneficial-caching-measure}) simplifies to \(\n \s{f} / \s{\rho}\), representing the intuitive notion that the benefit gained in generating \(\rho\) when \(f\) is cached is simply proportionate to the number of components no longer needing to be generated. Substituting this simplification and other uniform probabilities into Eq. (\ref{Eq:final-model}) (with normalizer), and rearranging, we obtain:
\begin{equation}
    \footnotesize
    p(c \mid f; \R) 
    \propto \mathop{size}(f)
    \sum_{\mathclap{\rho \in \bigcup_{x \in \mathcal{X}} \mathcal{B}_x}}
    \n \cdot w^{(\rho)}
\end{equation}
where \(w^{(\rho)} = 1 / \mathop{size}(\rho) |\mathcal{D}|^{\mathop{size}(\rho)}\).
The benefit of chunking a function for a uniform inference distribution is proportionate to a product of the function's size and a weighted count of where the function can be used. 
This product matches the high-level compression objectives used by other library learning approaches \cite{bowers2023top, ellis2021dreamcoder, cao2023babble, lazaro2019beyond}, balancing the two properties of a highly compressive function: one that is general enough to be used frequently, yet specific enough to capture lots of functionality when used.
In our case this balance arises organically when there is no knowledge to extract from the recognition model (it is uniform).

The weight associated with each function occurrence is inversely proportional to the size of the encompassing program: all else being equal, preference is given to functions within smaller programs over those in larger ones.
This property is desirable for compression in this context: larger programs offer more functions (flexibility) for later compression of it and future examples.




\section{Evaluation}
\label{sec:evaluation}

\begin{figure}[h]
    \centering
    \includegraphics[width=0.95\columnwidth]{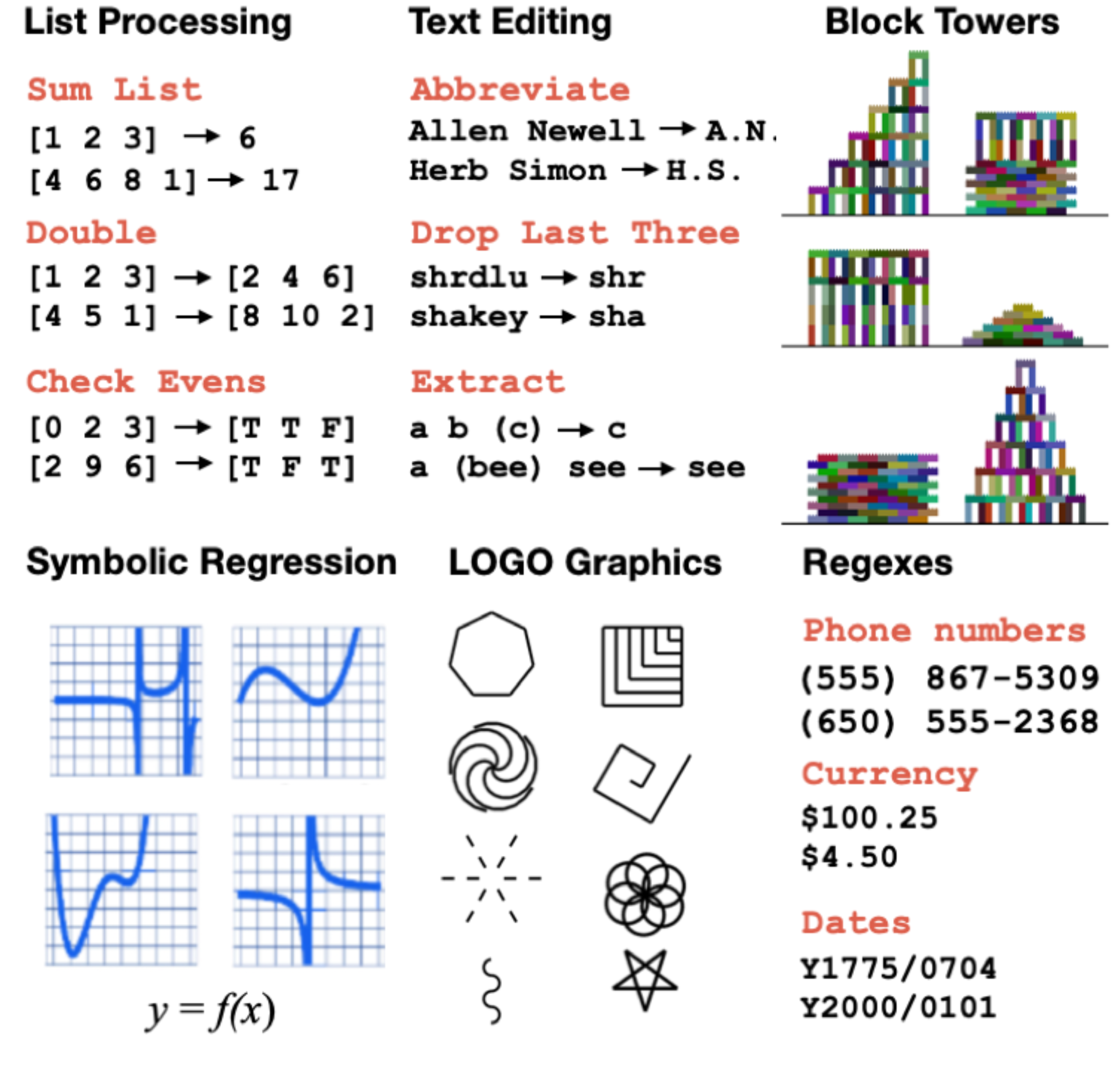}
    \caption{Example tasks from each domain tested as part of the evaluation. Figures taken from \cite{ellis2021dreamcoder}.}
    \label{Fig: example tasks}
\end{figure}

The aim of this paper is to introduce the alternative approach for library learning and demonstrate that the knowledge learnt by the recognition model for guiding program search space can be leveraged to directly influence how to restructure the search space, via chunking, for more effective inference.
We evaluate the two dream decompiler variants outlined in Section \ref{Section: Approach} to \textsc{DreamCoder} across 6 program synthesis domains. Each domain was used as part of the evaluation of \citet{ellis2021dreamcoder}, where \textsc{DreamCoder} was found to solve at least as many tasks as the best alternative tested for that domain and did so (mostly) in the least amount of time.
Below, we provide a summary of each domain. For example tasks, see Fig. \ref{Fig: example tasks}, and for further details, refer to \citet{ellis2018learning, ellis2021dreamcoder}.

\begin{compactenum}
    \item \textit{List processing}: synthesising programs to manipulate lists. The system is trained on 109 observed tasks and tested on 78 tasks. Each task contains 15 input/output examples. The initial library contains 
    common functional programming operations and elements.

    \item \textit{Text editing}: synthesising programs to edit strings of text. The system is trained on 128 randomly generated text editing tasks (each with 4 input/output examples) and tested on 108 tasks from the SyGuS competition \cite{alur2017sygus}. The initial library contains most programming operations from list processing, along with a component to represent all unknown string constants. 

    \item \textit{Block towers}: synthesising programs to build block towers. The system is trained on 54 observed tasks and tested on 53 held-out tasks. Tasks require constructing a target tower using an induced program that controls a simulated `hand' to move and drop blocks.

    \item \textit{Symbolic regression}: synthesising polynomial and rational functions. The system is trained and tested with 100 functions each. All polynomials have a maximum degree of 4. Tasks include 50 input/output pairs produced by the underlying function. The initial library consists of the \texttt{+}, \texttt{*}, and \texttt{÷} operators, along with a component for unknown real numbers. 

    \item \textit{LOGO graphics}: synthesising programs to draw images. The system is trained and tested on 80 tasks, where each is specified by a single image. The initial library contains primitives for constructing LOGO Turtle \cite{thornburg1983logo} graphic programs.

    \item \textit{Regexes}: synthesising probabilistic generative models. The system is trained and tested on 128 tasks each, originally sourced from \citet{hewitt2020learning}. Each task includes 5 strings, assumed to be generated by an unknown distribution that the system aims to infer as a regex probabilistic program.
\end{compactenum}


\begin{figure*}[!ht]
    \centering
    \includegraphics[width=0.95\textwidth]{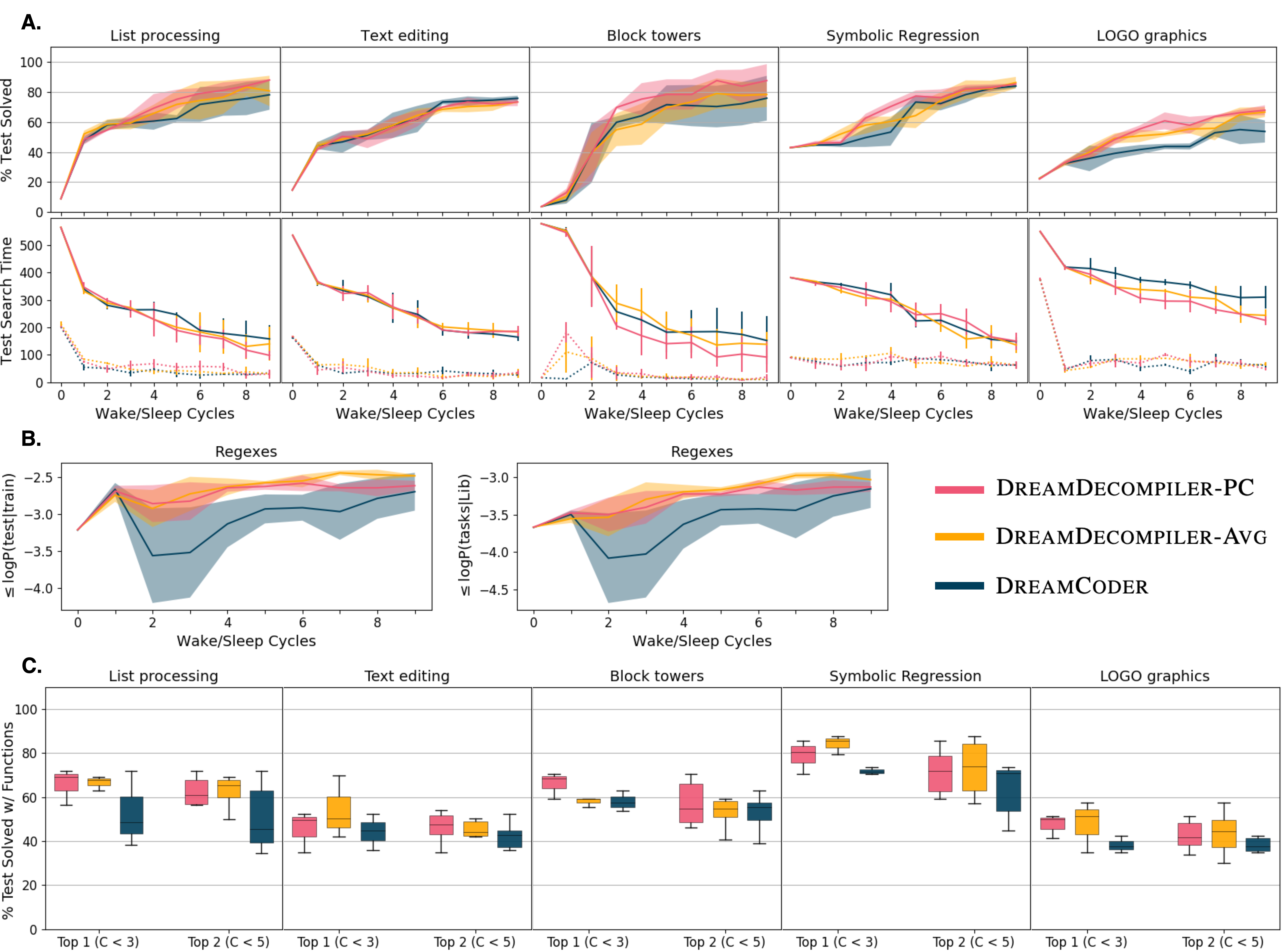}
    \caption{
    \textbf{(A)} Performance in deterministic program domains using the learnt library and recognition model of each wake-sleep cycle. Row 1 displays test set accuracy. Row 2 shows the average search time for test set solutions: solid lines represent the time averaged across all tasks, while dotted lines display the time averaged over solved tasks only. 
    \textbf{(B)} Performance with probabilistic programs. The left graph shows the marginal likelihood of strings in each task, given the learnt library. The right graph shows the posterior predictive probability of held-out strings, given the programs inferred from the task's observed strings.
    \textbf{(C)} Percentage of test tasks solved using the top 1 or top 2 functions chunked during the first 3 or 5 cycles (C) respectively. 
    All results are based on three random seeds with ±1 standard deviation. 
    }
    \label{Fig: Experiment graphs}
\end{figure*}

We use the same implementation\footnote{\url{https://github.com/ellisk42/ec}}, architecture and settings for the main \textsc{DreamCoder} model presented in \cite{ellis2021dreamcoder}, unless explicitly stated. 
The two dream decompiling variants are evaluated within the same \textsc{DreamCoder} system, with the \textit{only difference} being the code called to update the library. This allows us to isolate the effect of chunking choice: we leave broader system changes, which may enhance performance in conjunction with dream decompiling, for future work (see Section \ref{sec:future_work}).

The systems are evaluated on each domain for 10 of the wake-sleep cycles described in Section \ref{sec:dreamcoder}. In each cycle the system observes a random batch of training tasks during the wake-phase. The size of the batch and the time provided to search for solutions varies across domains, with smaller batch sizes and time limits used in most domains compared to \citet{ellis2021dreamcoder} (see Appendix \ref{A:hyperparameters}). Harder domains place greater emphasis on chunking choice and require less computational resources. 
Evaluation time varies across domain, with most taking roughly half a day using 1 NVIDIA A40 and 20 CPUs (40 for LOGO graphics). For regexes and LOGO graphics, evaluations extend to over a day.

Each cycle the system is additionally tested on the domain’s held-out tasks. Testing time is consistent across all domains: the system is provided 10 minutes per task to search for a solution using its current library and recognition model.
Fig. \ref{Fig: Experiment graphs}A (Row 1) shows the percentage of test tasks solved in each cycle by all systems. 
Except for text editing, where performance remains comparable across all systems, utilizing the recognition model for chunking (dream decompiling) enables faster domain proficiency and enhanced generalization through the learnt library.
This distinction is most evident during the intermediate cycles of learning, following similar performance in the initial iterations and before proceeding to converge again in the later iterations. Notably, this occurs despite all systems having solved a similar number of training tasks throughout (Fig. \ref{Fig:train_performance}, Appendix \ref{A:train_performance}). 
At the respective peak differences (excluding text editing), \textsc{DDC-PC} outperforms \textsc{DreamCoder} by \(13.25\%\) on average test task performance in list processing and by over \(17\%\) in the remaining domains. This is achieved despite having only solved \(3.7\%\) more training tasks at the time in list processing (cycle 5) and \(10.1\%\), \(3.1\%\), and \(2.1\%\) more training tasks at the time in block towers (cycle 7), symbolic regression (cycle 4), and LOGO graphics (cycle 5), respectively.

The larger generalization gaps suggest that more useful components, complementing the knowledge that is learnt by the recognition model to guide search, are being extracted from the same experience. The diminishing difference indicates that dream decompiling is particularly advantageous in scenarios where fewer domain solutions (data) are available to inform chunking decisions. This advantage arises by leveraging the compiled knowledge acquired by the recognition model---knowledge obtained not only from real data but also from fantasy data. 
A Welch's t-test can be performed for each cycle of learning to test the null hypothesis that \textsc{DDC-PC} and \textsc{DreamCoder} have equal average performance. We examine general performance across all domains by combining model samples from all domains and ensuring appropriate shifting to account for inter-domain differences in difficulty. This is achieved by subtracting the domain's average end performance with both models. The results show statistical significance in cycle 3 (\(t: 2.57,~p: 0.016\)), cycle 4 (\(t: 2.1,~p: 0.045\)), and the last few cycles, supporting claims of improved bootstrapping early on with better chunking decisions being made from the same experience.

To further test this claim, we examine the usefulness of functions chunked by each system in the earlier wake-sleep cycles. We quantify the usefulness of a function \(f\) as the percentage of test tasks eventually solved with a program containing \(f\) by the end of training. Fig. \ref{Fig: Experiment graphs}C (left grouping for each domain) shows the highest percentage of test tasks solved with a single function chunked by each system during the first 3 wake-sleep cycles. The right grouping shows the average percentage of test tasks solved with the top 2 functions chunked during the first 5 wake-sleep cycles. Across all domains, functions chunked early by the dream decompiling approaches are eventually utilized to solve more test tasks. This includes the text editing and symbolic regression domains, where each system solves an equal number of test tasks. Here, dream decompiling is still advantageous in making more beneficial single chunking decisions when example solutions are limited.



Fig. \ref{Fig: Experiment graphs}A (Row 2) shows average search time on held-out tasks, which decreases as learning progresses. By the end of learning, each system induces their program solutions (dotted lines) in near-identical average times across all domains. Notably, this holds true even in the list processing, block tower and LOGO graphic domains, where \textsc{DDC-PC} solves more end tasks; the combined knowledge learnt to simplify search enables access to larger relevant parts of program space \textit{without} compromising on time—which is reflected in the lower average search times on all tasks (solid lines). 

As discussed in Section \ref{Section: Approach}, \textsc{DDC-Avg} lacks the ability to determine how many functions should be chunked; it can only rank them. 
Although \textsc{DDC-PC} addressed this limitation, determining a precise threshold for chunking a specific function still relies on unknown cost/benefit payoffs, while also accounting for the lower bound approximation.
To evaluate both variants, we adopt a simple strategy of chunking the top \(k\) candidates each cycle, with \(k\) set to 2 for list processing, text editing and block towers, and 1 in the remaining domains.
This provides an initial investigation, highlighting the potential of the approach to enhance learning. 



\textbf{Size of chunked functions}~~
Across all deterministic domains, the library learnt by \textsc{DDC-PC} consistently outperforms \textsc{DDC-Avg}, validating the use of its extra complexity.
Additionally, in almost all domains the functions chunked by \textsc{DDC-Avg} are smaller in size than those chunked by \textsc{DDC-PC} (Table \ref{Table: average size}), reflecting the second issue identified with \textsc{DDC-Avg} in Section \ref{Section: Approach}.
%

\begin{table}[h]
  \caption{Average size of chunked functions in each domain.}
  \label{Table: average size}
  \centering
  \begin{tabular}{lcccccc} 
    \toprule
                       & LP     & TE & BT & SR & LG  & R\\
    \midrule
    \textsc{DDC-PC}    & 5.9 & 7.8 & 8.2 & 4.6 & 5.5 & 3.9 \\
    \textsc{DDC-Avg}   & 5.5 & 8.2 & 6.5 & 3.0 & 4.3 & 3.1 \\
    \textsc{DreamCoder}& 5.8 & 6.4 & 8.2 & 4.9 & 5.8 & 3.5 \\
    \bottomrule
  \end{tabular}
\end{table}

\section{Related Work}
\label{Section:related-work}
In this section we consider recent progress in library learning, focusing specifically on improvements proposed within, and in comparison to, the \textsc{DreamCoder} system.

Library learning requires addressing two subproblems: how to (i) \textit{generate} candidate functions for chunking, and (ii) \textit{select} amongst the candidates for useful additions to the library.
Recent advancements in candidate generation (subproblem i) are seen in \textsc{babble} \cite{cao2023babble} and \textsc{Stitch} \cite{bowers2023top}: \textsc{babble} with improved expressiveness—proposing common functionality despite syntactic differences (using domain-specific equational theories)—and \textsc{Stitch} with significantly improved efficiency—both time and memory use.
For candidate selection (subproblem ii), both approaches adopt the compression objective from \textsc{DreamCoder}—which is where we do something different, departing from compression in favour of chunking functions to complement the knowledge learnt to compose them.
Our focus here was solely on selection, assuming a readily available set of candidate functions. Enhanced candidate generation from either of these approaches could thus synergize well with dream decompiling.

\citet{wong2021leveraging} likewise introduce an alternative approach to candidate selection (subproblem ii) in library learning: an extension of \textsc{DreamCoder}'s compression objective that incorporates natural language annotations, modelling the hypothesis that humans often learn domain concepts with natural language descriptions.

\section{Conclusion}
\label{Section: Conclusion}

Becoming an expert in a program synthesis domain requires learning knowledge to alleviate the search problem. \textsc{DreamCoder} amortizes the cost of search by training a neural recognition model to reduce search breadth---effectively compiling useful information for composing program solutions across tasks. It reduces the depth of search by building a library capable of expressing discovered solutions in fewer components. As the library grows, a new recognition model is trained to exploit it and solve new tasks, thus bootstrapping the learning process.

Chunking program components should complement the knowledge that will be used to compose them. For this we introduced a novel approach to chunking that leverages the neural recognition model---effectively decompiling the knowledge learnt to generate program solutions across tasks and select high-level functions for the library. Consequently, the amortized knowledge learnt to reduce search breadth is now also used to reduce search depth. We show that this can lead to a stronger bootstrapping effect: better chunking decisions can be made from the same experience, leading to faster domain proficiency and improved generalization.

\section{Future Work and Limitations} 
\label{sec:future_work}
The experiments aimed to quantify the impact of dream decompiling by exclusively altering the selection of candidates generated by \textsc{DreamCoder}. The results provide support to pursue enhanced implementations of the theory presented, including promising system-wide changes.
As mentioned in Section \ref{Section:related-work}, candidate generation is one.
Functionality absent from existing program solutions are not generated by compression-focused approaches, yet may still score highly under dream decompiling (see Section \ref{sec:ddc}).
Additional approaches to candidate generation, like incorporating candidates sampled from the recognition model, are needed.

In the current system cycle (Fig. \ref{fig:chain-of-influence}), the library is updated after the wake-phase, where extra information, not utilized by the recognition model for chunking, may be available. Introducing a secondary training phase, occurring after waking but before updating the library, to ``compile in'' this newfound information could offer further improvements. 

Both variants of dream decompiling investigated can rank functions for chunking, but lack a clear means to determine an optimal number to chunk. While our experiments employed simple strategies to highlight the potential of their chunking choice, other strategies are likely more effective.


More sophisticated neural models remain an open direction for improving performance in \textsc{DreamCoder}'s framework. The enhanced amortization, achieved by reusing the neural model for chunking, implies that more sophisticated models could now have a cumulative impact on learning.

With dream decompiling, we are chunking functions that are useful for the \textit{current} recognition model to generate program solutions. However, in \textsc{DreamCoder}, once the library components used to specify the recognition model's distribution change, a new model is trained from scratch---with nothing learnt by the previous model passed on to the next. Exploring ways to continue learning \cite{kirkpatrick2017overcoming} when re-configuring its sample space could allow for improved, or at least more efficient, learning.

\section*{Acknowledgments}
We are grateful to Andrej Jovanović for helpful discussions, and to anonymous reviewers for their insightful suggestions, which have greatly improved the quality of the paper.

\section*{Impact Statement}
This paper presents work whose goal is to advance the field of Artificial Intelligence. There are many potential societal consequences of our work, none which we feel must be specifically highlighted here.


\bibliography{main}
\bibliographystyle{icml2024}

\newpage
\appendix
\onecolumn


\section{Handling more flexible recognition models}
\label{A:flexible_q}

Chunking with either \textsc{DreamDecompiler} variant requires calculating the probability of the recognition model generating a function as part of a program. In the main text we assumed, for simplicity, that the recognition model creates a distribution over programs by defining a distribution over library components and that the probability of the recognition model generating a program is equal to the product of independently generating each component of the program. In this case, calculating the probability of the recognition model generating a function (which is built from the same library of components) is straightforward. In this section we discuss how the probability of the recognition model generating a function can be calculated with the recognition model used by the \textsc{DreamCoder} \cite{ellis2021dreamcoder} system: a bigram distribution over well-typed programs. 

\subsection{Well-Typed programs}
\label{Types}
All library components have an associated type specifying the entities that their computations operate on and produce. This restricts the ways in which they can be combined to form valid programs. 

\textsc{DreamCoder}'s recognition model is constrained to be a distribution over standalone, well-typed programs of a \textit{requested} type only. Consequently, calculating the probability of a program being generated by the recognition model involves using a type inference algorithm. After supplying a program's type, the algorithm will track which components can be used at each point in the program's construction by determining which have a return type that unifies with what is required. To get the probability of seeing a program component, the recognition model's original distribution over all library components is renormalised to one over valid components (for this part of the program) only.

If a program contains a component whose return type does not unify with the required type, or if a component of a required type is expected but missing, then the program's probability is undefined. Therefore, to use \textsc{DreamCoder}'s type inference algorithm to calculate the probability of the recognition model generating a function requires that it is both a well-typed and a complete program. 
This is not the case for all the candidate functions in \(\F\) which are sub-expressions extracted from (refactored) programs. \textsc{DreamCoder} represents programs as polymorphic typed $\lambda$-calculus expressions. We can turn each candidate into a well-typed and complete program by performing $\eta$-conversion in reverse: wrapping each candidate function in an explicit function type (lambda abstraction) with as many variables as there are missing parts. By using a modified version of \textsc{DreamCoder}'s type inference algorithm that ignores the newly added variables, the probability of the recognition model generating the candidate function as part of a program can then be calculated. 

\subsection{Bigram model}
\label{Bigram Model}

In \textsc{DreamCoder}, the generative model's distribution over library components depends only on the requested type discussed in Section \ref{Types}: it is a \textit{unigram} distribution where the probability of generating a component is independent of any surrounding program context. On the other hand, the recognition model used by \textsc{DreamCoder} is a \textit{bigram} distribution over library components, conditioning on the function that the component will be passed to, and as which argument. 
Therefore, the probability of a (well-typed and complete) program $\rho$ with $K$ components being generated by the recognition model can be seen as a product of conditional probabilities $\prod_{k=1}^K q_{\R}(\rho_k \mid x, (\text{pa}_k, i_k, \tau_k))$, where $\text{pa}_k$ is component $k$'s parent, $i_k$ the argument index and $\tau_k$ the requested type. 


We are interested in calculating the probability of the recognition model generating a candidate function explicitly as \textit{part of} a program to solve a given task. With a unigram distribution, the recognition model could be used to score a candidate function (that has been turned into a well-typed program) directly. However, doing this with a bigram distribution would result in the probability of the recognition model generating the function as the \textit{outermost} top-level function of a solution only—i.e. when it has no parent.

To see why this is not desirable, consider a set of tasks requiring you to manipulate lists of integers. As the only programs solving these tasks are those returning lists of integers, the only top-level functions that the recognition model is learning to infer are those returning lists of integers. The probability of the recognition model generating a function that returns, say, a Boolean (such as indicating if an element of a list is positive) would be extremely low as the initial top-level function, even if its probability of being generated elsewhere is high. 

To find the likelihood \(q_{\R}(f \mid x)\) of the recognition model generating a function as part of a program to solve a task we must average over all the places and ways in which the function could be generated—i.e. over all possible $(\text{pa}, i, \tau)$ triplets. With a unigram distribution, there is no difference in where a function is generated (other than the type constraints)—and thus nothing to average over. For a bigram distribution, the same is true for all function components other than the root: once the root has been generated, the probability of the rest are conditionally independent to where in a program they are being generated. Therefore, by splitting our function into its root component $f_r$ and remaining components $f_{\setminus r}$, our desired probability \(q_{\R}(f \mid x)\) equals:
\begin{equation}
    \mathbb E_{p((\text{pa}_r, i_r, \tau_r) \mid x)}[q_{\R}(f_{\setminus r} \mid x, f_r)q_{\R}(f_{r} \mid x, (\text{pa}_r, i_r, \tau_r))]
    \label{Eq: context marginal}
\end{equation}

Calculating the expectations requires specifying the probability distribution \(p((\text{pa}_r, i_r, \tau_r) \mid x)\) over the context in which a function could be generated. 
One option is to place a uniform distribution over each option that could conceivably arise: every argument of every component in the library currently being used to generate programs (along with their respective types), has equal probability. 
Though simple, this option has two downsides.
First, one aspect motivating the use of the recognition model for chunking was to remove the human defined priors used by \textsc{DreamCoder}. This option we would be doing the exact same. 
The second disadvantage has to do with the distribution itself: a library that contains only a few components that expect a certain type implies that all functions returning that type would be disfavoured. However, even if we expect the library to reflect some information about the tasks it is used to solve, a few library components requesting a certain type should not imply that the likelihood of generating functions returning that type should be low: those few components could be used in multiple ways.

One might be tempted, then, to use the recognition model's distribution over parent components, but this would itself require a prior over $(\text{pa}, i, \tau)$ triplets, leading to an infinite regress. Instead, we can achieve a similar effect by using a distribution over $(\text{pa}, i, \tau)$ triplets proportionate to their frequency in the program solutions that the recognition model helped discover during search. Note that the distribution in Eq. (\ref{Eq: context marginal}) is conditioned on a specific task. Therefore, it may be more appropriate to only use the programs found to solve that specific task (i.e. those in $\mathcal B_x$). However, a small beam size or even just low (or empty) program beams early on in learning would not provide a meaningful distribution.

\section{Hyperparameters}
\label{A:hyperparameters}
The same base \textsc{DreamCoder} system and hyperparameter values are used to compare the different approaches to chunking. The final systems differ only in the method used to update the library. In our experiments, almost all hyperparameters are set to the same value as those used by the main \textsc{DreamCoder} model presented in \citet{ellis2021dreamcoder}. 
Table \ref{Table: hyperparameters} shows the hyperparameter values (relevant to all systems) that differ in at least one domain compared to those used in \citet{ellis2021dreamcoder}. 
Additionally, \textsc{DreamCoder} employs a hyperparameter, denoted as \(\lambda\) in \citet{ellis2021dreamcoder}, which controls the prior distribution over libraries. In their work, \(\lambda\) was consistently set to 1.5 for all domains, except for symbolic regression where it was set to 1. In our experiments, we maintain uniformity by using the same \textsc{DreamCoder} model, setting \(\lambda\) to 1.5 for all domains.

\begin{table}[h]
  \centering
  \begin{tabular}{lcccc}
    \toprule
      Domain   & Wake timeout (m) & Batch size & CPUs \\
    \midrule
    List Processing     & 12  & 10 & 20 \\
    Text Editing        & 12 & 10 & 20 \\
    Block Towers        & 5 & 15 & 20 \\
    Symbolic Regression & 2 & 10 & 20 \\
    LOGO Graphics       & 12 & 20 & 40 \\
    Regexes            & 6 & 10 & 20 \\
    \bottomrule
  \end{tabular}
  \caption{Hyperparameters used for experiments that differ in at least one domain from those used by the main \textsc{DreamCoder} model presented in \cite{ellis2021dreamcoder}, where all other hyperparameters can be found.}
  \label{Table: hyperparameters}
\end{table}



\section{Evaluation Training Performance}
\label{A:train_performance}
\begin{figure*}[h]
    \centering
    \includegraphics[width=0.95\textwidth]{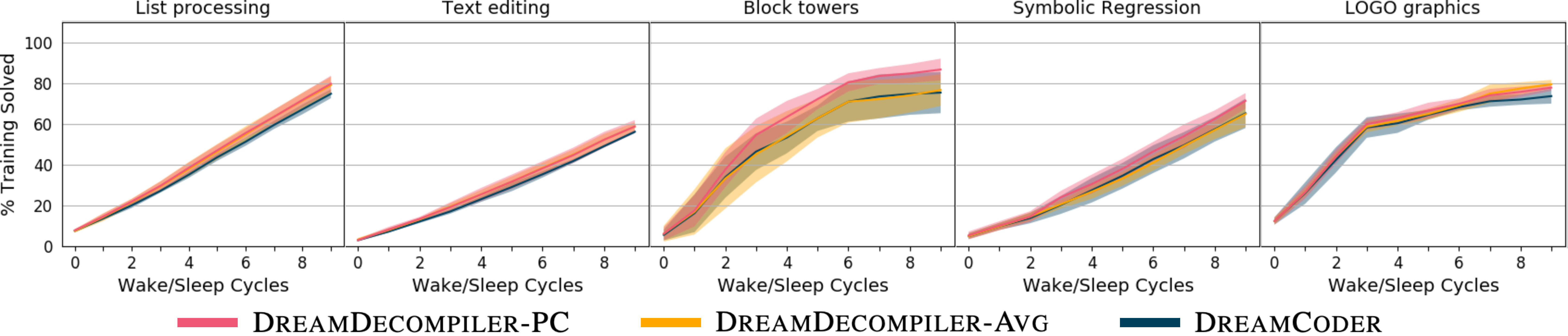}
    \caption{
    Performance on training tasks in deterministic program domains using the learnt library and recognition model of each wake-sleep cycle.
    All results are based on three random seeds with ±1 standard deviation. 
    }
    \label{Fig:train_performance}
\end{figure*}

Fig. \ref{Fig:train_performance} shows the percentage of train training tasks solved by each system for the same wake-sleep cycles shown in Fig. \ref{Fig: Experiment graphs} in Section \ref{sec:evaluation}. Recall that as part of the \textsc{DreamCoder} system (which all tested library learning approaches inherit and share), only a small batch of training tasks is provided for solving within the wake timeout during each cycle (see Table \ref{Table: hyperparameters} for specific values). This feature imposes a limit on the improvement that can be made from one cycle to the next, in contrast to test performance where the system attempts to solve as many held-out tasks as possible. The key takeaway from these graphs is the following: although all systems solve a similar number of training examples throughout and thus have access to the same number of training solutions for learning, the dream decompiling approaches consistently demonstrate greater generalization on the held-out tasks using the library learnt from these examples, as seen in Fig. \ref{Fig: Experiment graphs} in Section \ref{sec:evaluation}.

\end{document}